\documentclass[11pt]{article}
\usepackage{geometry}
\usepackage{coling2020}
\usepackage{times}
\usepackage{url}
\usepackage{latexsym}
\usepackage{microtype}
\usepackage{graphicx}
\usepackage{subcaption}
\usepackage{amsmath}
\usepackage{capt-of}
\usepackage{hhline}
\colingfinalcopy 

\title{ERNIE at SemEval-2020 Task 10: Learning Word Emphasis  Selection by Pre-trained Language Model}

\author{Zhengjie Huang, Shikun Feng, Weiyue Su, Xuyi Chen, \\ 
  \textbf{Shuohuan Wang, Jiaxiang Liu, Xuan Ouyang, Yu Sun} \\
  Baidu Inc., China \\
  {\tt \{huangzhengjie,fengshikun01,suweiyue,chenxuyi\}@baidu.com } \\
  {\tt \{wangshuohuan,liujiaxiang,ouyangxuan,sunyu02\}@baidu.com } \\}

\date{}

\begin{document}
\maketitle
\begin{abstract}
  This paper describes the system designed by ERNIE Team which achieved the first place in SemEval-2020 Task 10: Emphasis Selection For Written Text in Visual Media. Given a sentence, we are asked to find out the most important words as the suggestion for automated design. We leverage the unsupervised pre-training model and finetune these models on our task. After our investigation, we found that the following models achieved an excellent performance in this task: ERNIE 2.0, XLM-ROBERTA, ROBERTA and ALBERT. We combine a pointwise regression loss and a pairwise ranking loss which is more close to the final $Match_{m}$ metric to finetune our models. And we also find that additional feature engineering and data augmentation can help improve the performance. Our best model achieves the highest score of 0.823 and ranks first for all kinds of metrics.
\end{abstract}

\section{Introduction}
\label{sec:introduction}
\blfootnote{
    %
    %
   
    %
    %
    %
    %
     \hspace{-0.65cm}  
     This work is licensed under a Creative Commons 
     Attribution 4.0 International License.
     License details:
     \url{http://creativecommons.org/licenses/by/4.0/}.
}

Emphasis selection for written text in visual media is proposed by \newcite{shirani2020semeval} and \newcite{shirani-etal-2019-learning}. The purpose of this shared task is to design automatic methods for emphasis selection, i.e. choosing candidates for emphasis in short written text, to enable automated design assistance in authoring. For example, \newcite{shirani-etal-2019-learning} mentions that such a technique can be applied to some graphic design applications such as Adobe Spark to perform automatic text layout using
templates that include images and text with different fonts and colors. The major challenge is that given only thousands of annotated short text data without any context about the text or visual background images, we are asked to learn the author- or domain-specific emphatic about the short text. Besides, these short text data are annotated by crowd-sourcing workers. And we find that different annotators have different standards, which increases the difficulty of this task.

To identify the most important words, we model the task as a sequential labeling problem. Our base models leverage different unsupervised language model such as ERNIE 2.0 \cite{sun2019ernie}, XLM-ROBERTA \cite{conneau2019unsupervised}, ROBERTA \cite{liu2019roberta} and ALBERT \cite{lan2019albert}. These large unsupervised models are pre-trained on a large amount of unannotated data and carry valuable lexical, syntactic, and semantic information in training corpora. Our approach is as follows: firstly, the word-level output representations for the sentence are computed by pre-trained models and then fed into a designed downstream neural network for word selections; secondly, we finetune the downstream networks together with the pre-trained models on the annotated training data; thirdly, we investigate several different objective functions to learn our model; and finally, we apply feature engineering and several data augmentation strategies for further improvement.

The rest of the paper is organized as follows. In Section \ref{sec:overview}, we will briefly overview some related works to our system. Section \ref{sec:approach} shows the details of our approach. Our experiments will be shown in Section \ref{sec:exp}, and Section \ref{sec:clu} concludes.

\section{Related Work}
\label{sec:overview}

Recently pre-trained models have achieved state-of-the-art results in various language understanding tasks such as BERT \cite{devlin2018bert}, XLM-ROBERTA \cite{conneau2019unsupervised}, ROBERTA \cite{liu2019roberta}, ALBERT \cite{lan2019albert} and ERNIE 2.0 \cite{sun2019ernie}. \cite{devlin2018bert} first introduced a bidirectional encoder representation from transformers called BERT. They developed several pretraining strategies such as masked language models and a next sentence prediction task. Since then, many studies about pretraining strategies come out and most of them share similar neural network architectures but with different pretraining schemes. For example, ROBERTA \cite{liu2019roberta} finds that dynamically changing the masking for the masked language model and removing the next sentence prediction task can help with improving the performance in downstream tasks. ALBERT \cite{lan2019albert} adds sentence order prediction task and optimizes the memory usage of original BERT to achieve better results. XLM-ROBERTA \cite{conneau2019unsupervised} trains its models over one hundred languages with multilingual settings and gets the first single large model for all languages.

ERNIE 2.0 \cite{sun2019ernie} is an improvement of ERNIE 1.0 \cite{sun2019ernie1} and the world’s first model to score over 90 in terms of the macro-average score on GLUE benchmark \cite{wang2018glue}.  ERNIE 1.0 \cite{sun2019ernie1} introduces knowledge masking strategies. It gains a large benefit from entity-level and phrase-level masked language models. ERNIE 2.0 \cite{sun2019ernie} proposes a continual pre-training framework that incrementally builds pre-training tasks and then
learns pre-trained models on these constructed tasks via continual multi-task learning. ERNIE 2.0 constructs three kinds of tasks including word-aware tasks, structure-aware tasks and semantic-aware tasks. All of these pre-training tasks rely on self-supervised or weak-supervised
signals that could be obtained from massive data without human annotation. A continual multi-task learning method is proposed to improve the model's memory over different pre-training tasks. 

The researchers of ERNIE 2.0 released a new version recently which made a few improvements on knowledge masking and application-oriented tasks, aiming to advance the model's general semantic representation capability. In order to improve the knowledge masking strategy, they proposed a new mutual information-based dynamic knowledge masking algorithm. They also constructed specific pre-training tasks for different applications. For example, they added a coreference resolution task to identify all expressions in a text that refer to the same entity. Details can be found on the blog\footnote{http://research.baidu.com/Blog/index-view?id=128} .  


Following the current trends of pre-training and fine-tuning paradigm for natural language processing, our system adopts these models as our base word and sentence representation.

\section{Our Approach}
\label{sec:approach}

\subsection{Word Emphasis Regression with Subword Alignment}

\begin{figure}[htbp]
	\centering
	\begin{subfigure}[b]{.45\textwidth}
	\includegraphics[scale=0.5, clip]{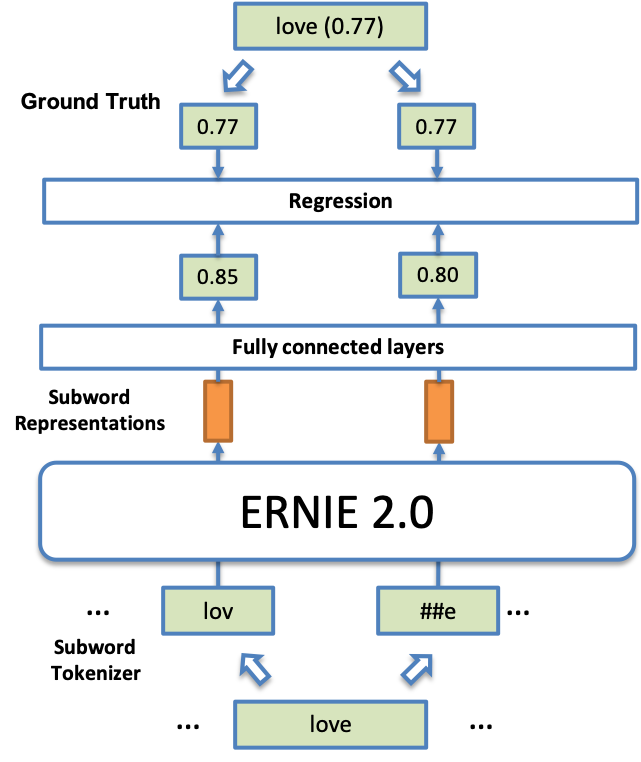}
	\caption{}
	\label{fig:regression_model_2}
    \end{subfigure}
	\begin{subfigure}[b]{.45\textwidth}
	\includegraphics[scale=0.5,clip]{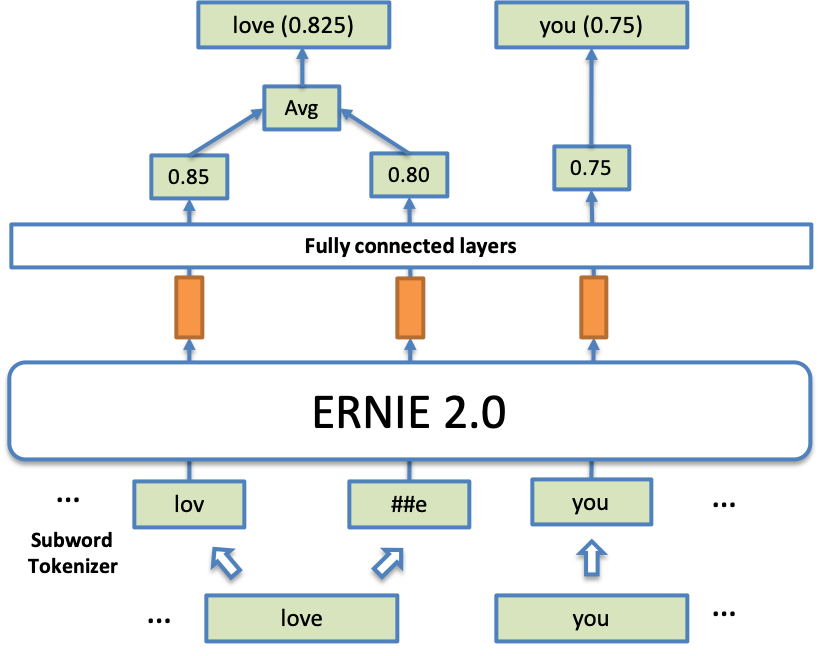}
	\caption{}
	\label{fig:regression_model_3}
    \end{subfigure}
    \caption{The input sentence will be tokenized into subword level by WordPiece algorithm. Then we use pre-trained models such as ERNIE 2.0 to get the subword representations. Next we apply a regression layer on the model as described in Figure \ref{fig:regression_model_2}. Finally we will aggregate subword scores to get the final word emphasis like Figure \ref{fig:regression_model_3}.  }
    \label{fig:regression_model}
\end{figure}

Instead of learning label distributions with KL divergence loss like \newcite{shirani-etal-2019-learning}, we directly learn to regress the emphasis probability values with mean squared error (MSE) loss. Our model is shown in Figure \ref{fig:regression_model}. We simply plug in the task-specific inputs and outputs into the pre-trained model like ERNIE 2.0 does. Words are preprocessed into subword level with the WordPiece tokenizer \cite{wu2016google} like most of the BERT-style fine-tuning tasks. The subword tokens are then fed into ERNIE 2.0 to compute the contextual representations for each subword. The task-specific output layer is a fully connected neural network with sigmoid activation to constrain the output between 0 and 1. Since our model is based on the subword level, the ground-truth scores of the words are split into pieces and each subword piece will learn its aligned emphasis value. All the parameters of ERNIE 2.0 and the final fully connected neural network will be tuned together. During the inference stage, word emphasis score will be computed by aggregating its corresponding subword scores on average as shown in Figure \ref{fig:regression_model_3}.

\subsection{Subword-level Pairwise Ranking Loss}

\begin{figure}[htbp]
    \centering
    \includegraphics[scale=0.60, trim=0 0cm 0 0cm,clip]{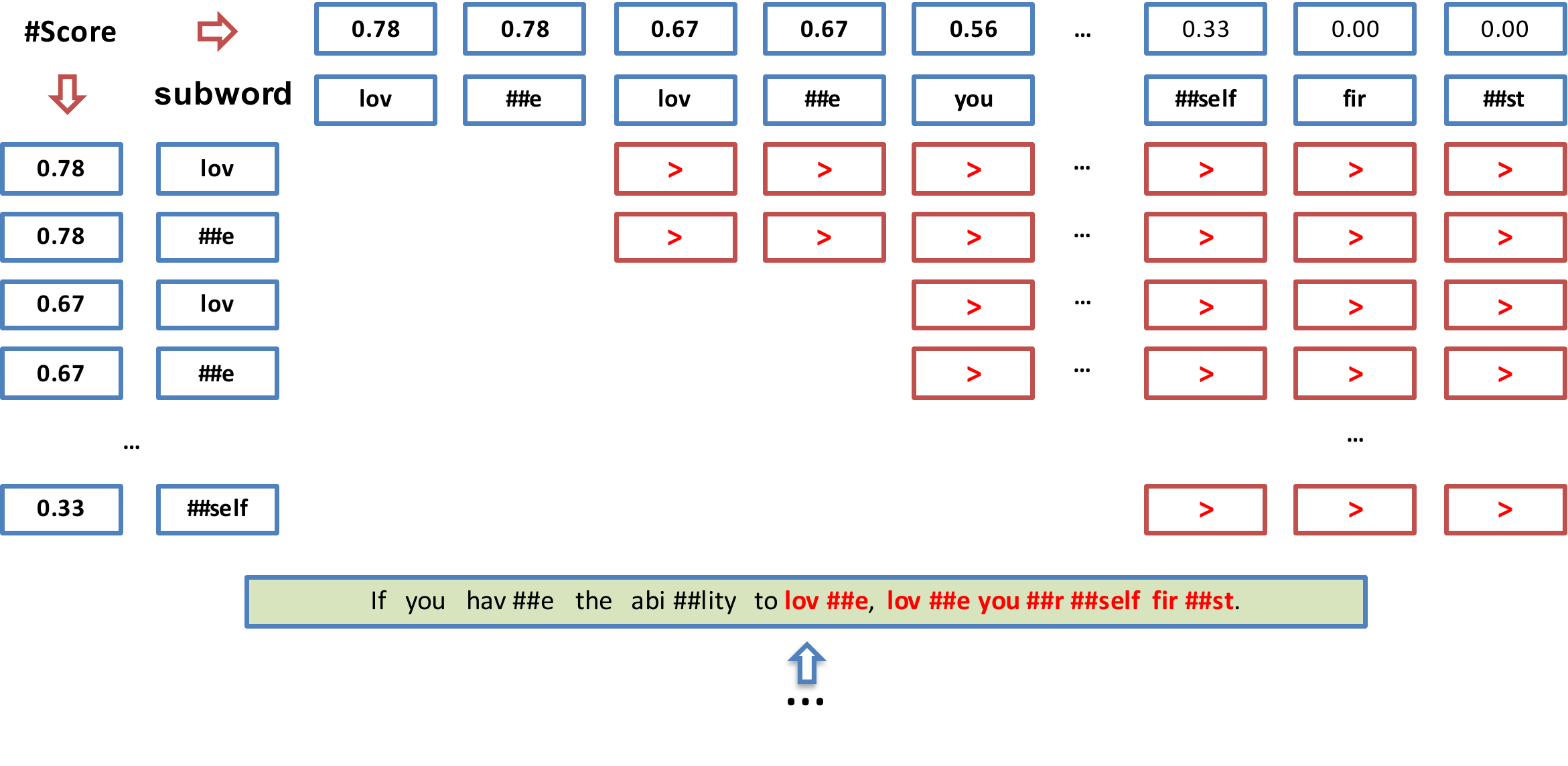}
	\caption{Pairwise ranking loss.}
	\label{fig:pairwise-ranking-loss}
\end{figure}

Since the final metrics only consider the top 4 words with the highest emphasis scores, the individual subword level regression task ignores relative scores between the tokens and might hurt the performance. To overcome this issue, we develop a pairwise ranking loss, which considers all pairs of the subword pieces and learns the relative orders of the emphasis probability. As shown in Figure \ref{fig:pairwise-ranking-loss}, the emphasis will be also split at subword level. Each subword piece will be compared with all other subword pieces that have lower scores. Then the loss is computed as follows:

\begin{equation}
J = \frac{1}{N^2}\sum_{i}^{N}\sum_{j}^{N}- \max(score(w_i) - score(w_j), 0) * \log \sigma (s(w_i) - s(w_j)) ,
\end{equation} 
where $\sigma$ is sigmoid function and the  $score(.)$ function returns the ground-truth emphasis probability labels for each subword and the $s(.)$ is the output of the logits score (without sigmoid activation). The loss is weighted by the gap of scores.

\subsection{Word Emphasis Lexical Feature}
After further investigating the short text data, we find that some additional features about the capitalization of words and the appearance of hashtags can bring further improvements. The statistics of the average score for different word types are shown in Table \ref{tab:word_features}. Obviously, words with hashtag or capital letters are more likely to be annotated as important words. However, WordPiece algorithm separates words into pieces and drops the information about the prefix of words. For example, "\#plantgang" is split into "\#", "plant", and "\#\#gang". In our model, regression loss is computed for each individual word so it is difficult for the word piece "plant" and "\#\#gang" to capture the prefix information. Therefore, the explicit features about the special meaning of hashtag in social media and visual impact of uppercase characters are valuable. These features about the words are denoted as 0-1 vectors and concatenated with the ERNIE embedding as the inputs of the final fully-connected layers as shown in Figure \ref{fig:feature_eng}.

\begin{table}[htbp]
    \begin{minipage}[b]{0.56\linewidth}
    \centering
	\begin{tabular}{cc}
	    \hline
		\textbf{Word Types} & \textbf{Avg. Score} \\ \hhline{==}
		  All        & 0.284   \\
	      Starts with a capital letter & 0.369  \\
		Word in uppercase &  0.333 \\ 
		Starts with hashtag & 0.611 \\ \hline
	\end{tabular}
    \caption{Word types and its corresponding scores.}
    \label{tab:word_features}
    \end{minipage}
    \begin{minipage}[b]{0.4\linewidth}
    \centering
    \centering
    	\includegraphics[scale=0.4, trim=0 0cm 0 0cm,clip]{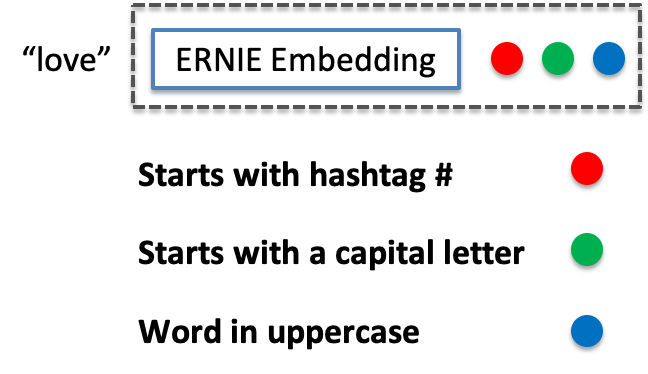}
        \captionof{figure}{Additional features.}
    	\label{fig:feature_eng}
    \end{minipage}

\end{table}

\subsection{Data Augmentations}
Because there are only 2000 annotated data, it's quite easy to overfit the training data even with pre-trained models. To enlarge the amount of annotated data, we design several data augmentation strategies as follows: 1) randomly remove a word, 2) randomly uppercase a word, and 3) randomly reverse a sentence. We find that the augmentation can help delay the overfitting occurrence, especially for large models. The details of the augmentation schemes are shown in Table \ref{tab:data_augmentation_setting}. Each scheme is triggered independently for each word and sentence with the given probabilities. Therefore, we can have many different modified versions of the origin sentences to delay the phenomenon of overfitting.

\begin{table}[htbp]
    \centering
	\begin{tabular}{cc}
	    \hline
		\textbf{Augmentation Schemes} & \textbf{Probability} \\ \hhline{==}
		  Randomly remove a word        & 1\%   \\
		Randomly uppercase a word & 5\%   \\
		Reverse the sentence &  10\%  \\ \hline
	\end{tabular}
    \caption{Data augmentation settings.}
    \label{tab:data_augmentation_setting}
\end{table}

\section{Experimental Results}
\label{sec:exp}

All experiments are executed on an Nvidia V100 GPU. Each model runs 10 epochs with early stopping strategies based on the performance in the validation set. Since the training and validation data are both small, we find that models have a large variant performances among each run.\footnote{The performance on different fold can have a large variance differents. For example, XLM-ROBERTA-LARGE can achieve rank score from 76.7\% to 80.1\% in different fold runs.} So we combine all the provided training and validation sets, then split them in a random 8-Fold settings. For each model in each fold, we run five times and we will report the average score on our 8-Fold settings to get a much more stable analysis. 

\begin{table}[htbp]
    \centering
	\begin{tabular}{cc}
	    \hline
		\textbf{Model} & \textbf{8FoldsCV-AVG. Rank} \\ \hhline{==}
		\textbf{ERNIE 2.0-LARGE}          & \textbf{0.7810}   \\
		XLM-ROBERTA-LARGE & 0.7791   \\
		ROBERTA-LARGE     & 0.7764   \\
		ALBERT-XXLARGE-v2 & 0.7755  \\ \hline
	\end{tabular}
    \caption{The results of different models.}
    \label{tab:models_scores}
  
\end{table}
\begin{table}
    \centering
	\begin{tabular}{ccc}
	    \hline
		\textbf{Strategies} & \textbf{Average Gain} & \textbf{Max Gain}     \\ \hhline{===}
		PairWise Loss     & -0.00544                  & 0.00385 \\
		\textbf{Lexical Features}  & \textbf{0.00105}                   & \textbf{0.00709} \\
		Data Augmentation & -0.00144                  & 0.00344 \\ \hline
	\end{tabular}
	\caption{The score gains from different strategies.}
	\label{tab:gains_strategies}
\end{table}

Table \ref{tab:models_scores} shows the score across different models, we find that ERNIE 2.0 is the most powerful base model among several different pre-trained models, and gets 0.781 average ranking score over 8-Fold cross-validation. 

\begin{figure}[htbp]
	\centering
	\includegraphics[scale=0.5]{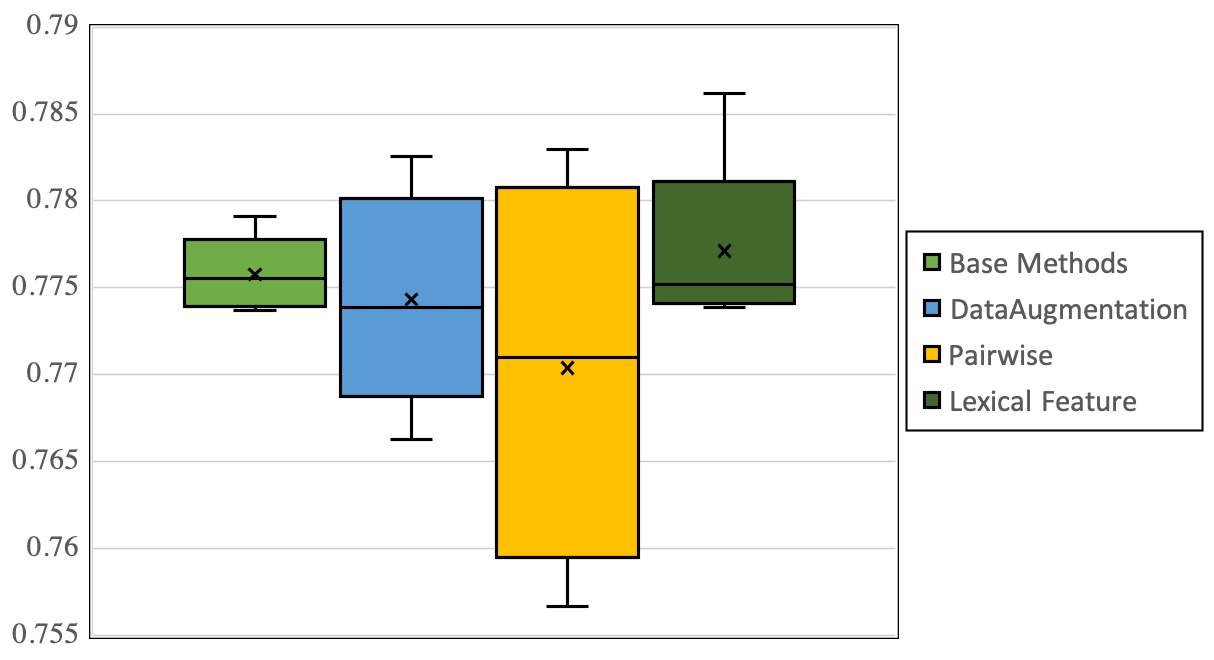}
	\caption{Validation scores among different strategies.}
	\label{fig:ablation_studies}
\end{figure}

Table \ref{tab:gains_strategies} shows the score gains across different fine-tuning strategies. We report the average gain and the maximum gain over the average score for each base model on the 8-Fold settings. We find that not all models can benefit from these training schemes. However, they still bring large improvement to some of our models. In Figure \ref{fig:ablation_studies}, we draw the box plot of the model scores. The box plot shows that the lexical feature is the most effective strategy. Besides, data augmentation and pairwise loss can also achieve a higher score. 

For final submission, we ensemble our best strategies and we achieve the highest score 0.823 ranking first for all kinds of the metrics.
\section{Conclusion}
\label{sec:clu}
In this paper, we present our system that ranks first in SemEval-2020 Task 10. Our solution contains several strategies and we provide detailed experiments to analyze which of them are effective. Our experiments show that models empowered by pre-trained language models are most effective, especially for ERNIE 2.0. Besides, lexical features, pairwise loss, and data augmentation can also bring improvement for some of our models.

\newpage
\bibliographystyle{coling}
\bibliography{semeval2020}

\end{document}